\documentclass[hyperref]{article}
\usepackage{tex/sty/packages}
\usepackage[accepted]{tex/sty/icml2022}
\usepackage{tex/sty/math}
\usepackage{tex/sty/colors}
\usepackage{tex/sty/commands}

\begin{document}

\icmltitlerunning{Learning to Cut by Looking Ahead}

\twocolumn[
\icmltitle{Learning to Cut by Looking Ahead: \\Cutting Plane Selection via Imitation Learning}



\icmlsetsymbol{equal}{*}

\begin{icmlauthorlist}
\icmlauthor{Max B. Paulus}{eth}
\icmlauthor{Giulia Zarpellon}{vector}
\icmlauthor{Andreas Krause}{eth}
\icmlauthor{Laurent Charlin}{hec,mila}
\icmlauthor{Chris J. Maddison}{uoft}
\end{icmlauthorlist}

\icmlaffiliation{eth}{Department of Computer Science, ETH Zürich, Zürich, Switzerland}
\icmlaffiliation{vector}{Vector Institute, Toronto, Canada}
\icmlaffiliation{hec}{Department of Decision Sciences, HEC Montréal, Montréal, Canada}
\icmlaffiliation{mila}{Mila, Montréal, Canada}
\icmlaffiliation{uoft}{Department of Computer Science and Department of Statistical Sciences, University of Toronto, Toronto, Canada}

\icmlcorrespondingauthor{Max Paulus}{max.paulus@inf.ethz.ch}
\icmlcorrespondingauthor{Giulia Zarpellon}{gzarpellon@cs.toronto.edu}

\icmlkeywords{Machine Learning, ICML, Integer Programming, Cutting Plane, Imitation Learning}
\vskip 0.3in
]



\printAffiliationsAndNotice{}  
\begin{abstract}

Cutting planes are essential for solving mixed-integer linear problems (MILPs), because they facilitate bound improvements on the optimal solution value. For selecting cuts, modern solvers rely on manually designed heuristics that are tuned to gauge the potential effectiveness of cuts. We show that a greedy selection rule explicitly \emph{looking ahead} to select cuts that yield the best bound improvement delivers strong decisions for cut selection---but is too expensive to be deployed in practice. In response, we propose a new neural architecture (NeuralCut) for imitation learning on the lookahead expert. Our model outperforms standard baselines for cut selection on several synthetic MILP benchmarks. Experiments with a B\&C solver for neural network verification further validate our approach, and exhibit the potential of learning methods in this setting.

\end{abstract}
\section{Introduction}
\label{sec:intro}

Mixed-Integer Linear Programming (MILP) problems are optimization problems in which some decision variables represent indivisible choices and are thus required to assume integer values. MILP models are used in many industrial contexts (e.g., logistics, production planning), and can also be solved to verify robustness of neural networks (NN) \cite{ChengEtAl17,TjengEtAl18}.
We write a MILP as 
\begin{equation}\label{eq:milp}
    z^{\mathit{OPT}} = \min\{c^T x: Ax\le b, x_j\in\mathbb{Z} \:\forall j\in I\},
\end{equation}
where $A\in\mathbb{R}^{m\times n}$, $b\in\mathbb{R}^m$, $c, x\in\mathbb{R}^n$, and $I\subseteq\{1, \dots, n\}$ defines which variables are required to be integral. 
MILPs are in general $\mathcal{NP}$-hard, but modern solvers effectively tackle them by means of a collection of both exact and heuristic techniques. In particular, the MILP algorithmic framework relies on repeatedly solving (and improving) relaxed versions of the original problem that are computationally tractable. When integrality requirements $x_j\in\mathbb{Z} \:\forall j\in I$ are dropped, one obtains the \emph{continuous relaxation} of \eqref{eq:milp}
\begin{equation}\label{eq:relax}
    z^* = \min\{c^T x : Ax\le b, x\in\mathbb{R}^n\},
\end{equation}
which is a linear problem (LP). Clearly, $z^* \le z^{\mathit{OPT}}$, and tightening the relaxation leads to better bounds on the optimal value of \eqref{eq:milp}, which can be used, e.g., to improve the efficiency of a Branch-and-Bound (B\&B) tree search procedure \cite{LandD60}---the typical framework used to solve MILPs. 

Tightening a MILP continuous relaxation can be done by introducing \emph{cutting planes} \cite{Gomory60} in the formulation.
Given an optimal solution $x^*$ for the LP relaxation \eqref{eq:relax}, a cut $C$ is a linear inequality $\pi^T x \le \pi_0$ that \emph{separates} $x^*$ from the convex hull of integer feasible solutions of \eqref{eq:milp}. Formally, $(\pi, \pi_0)\in\mathbb{R}^{n+1}$ is such that $\pi^T x^* > \pi_0$ and $\pi^T x \le \pi_0$ is valid for the feasible solutions of \eqref{eq:milp}.

Cutting planes are an essential component of the MILP resolution process, with many families of both general-purpose and class specific cuts theoretically studied and implemented in modern solvers \cite{DeyM18}. Still, their practical management is governed by manually-designed heuristic rules that decide, e.g., how many cuts should be added to the formulation from a pool of candidates, which types should be preferred, how many times the separation procedure should be repeated, and also which cuts should be removed in later optimization stages.
In particular, the process of selecting cutting planes from a pool relies on simple formulas that are tuned to gauge the potential effectiveness of cuts in solving MILPs, for example by measuring the extent by which a cut is violated by $x^*$ or its nonzero support.

A natural way of assessing the impact of a cut is to quantify its resulting bound improvement. With this motivation, we design a greedy \emph{lookahead} selection rule, explicitly trying cuts beforehand to select those yielding the largest gain in terms of LP bound. We show that following such criterion allows us to close the integrality gap more effectively than common heuristics, delivering strong decisions for cut selection (Section~\ref{sec:LAdesign}). However, looking ahead comes with too high of a computational overhead to be viable in practice. In response, we use the lookahead rule as an \emph{expert} to train NN policies via imitation learning; once trained, a single learned policy can be deployed on several MILP instances of the same family, so machine learning (ML) offers a way to amortize the cost of computing expert decisions.

Interfacing the MILP solver SCIP \cite{scip}, we focus on the core task of selecting a single cut from a pool of available ones. We extract useful information from the solver to populate with novel hand-crafted input features a tripartite graph encoding of the cut selection system. We then develop NeuralCut, a NN architecture that combines graph convolutions with attention to train cut selection policies through imitation of the lookahead rule (Section~\ref{sec:imitation}).  
Experiments on four synthetic families of MILPs confirm the ability of NeuralCut policies to mimic the expert, selecting bound-improving cutting planes better than standard heuristics and SCIP's own cut selection criterion. Trained policies are also successful in closing the integer gap when rolled-out on never-seen test instances (Section~\ref{sec:tangexp}).
Finally, we stress-test our framework on the challenging benchmark of NN Verification models \cite{NairEtAl20} and deploy NeuralCut in a realistic B\&C framework (Section~\ref{sec:spotlight}). Our results highlight the potential for improving solver performance with learned models for cut selection.

The present work shows that imitation learning is a viable approach to tackle and improve cutting planes selection in MILP solvers. Specifically, our contributions can be summarized in the following points.
\begin{itemize}
    \item We propose a lookahead criterion for cut selection that chooses cuts based on explicit computations of LP bound improvement, and use it as the expert in imitation learning experiments.
    \item We design a tripartite encoding for the cut selection system and develop NeuralCut, a novel NN architecture able to scale to multiple families of MILPs.
    \item We highlight NN Verification as a challenging benchmark for cut selection, exhibiting the potential of ML methods in this setting.
\end{itemize}

\section{Background}
\label{sec:background}

Cutting planes in MILP solvers are separated in successive \emph{rounds}. At a basic level, each round $k$ involves (i) solving a MILP continuous relaxation; (ii) inspecting a pool $\mathscr{C}^k$ of available cuts generated by the MILP solver 
to select a subset $S^k\subseteq\mathscr{C}^k$; (iii) adding $S^k$ to the continuous relaxation, before proceeding to round $k+1$.
On the one hand, incorporating all the cuts available in the pool into the formulation would maximally improve the bound and tighten the relaxation at each round. On the other hand, the addition of too many cuts would result in large models, which can become slower to solve and present numerical instabilities---so cut selection is necessary.

After $k$ separation rounds the LP relaxation includes all the original constraints $Ax\le b$, together with cutting planes $(\pi, \pi_0)$ selected along the way: $z^{k} = \min\{c^T x : x\in P^k\}$ is solved at round $k+1$, where 
\begin{equation}\label{eq:krounds}
    \begin{aligned}
    P^k &= \{x\in\mathbb{R}^n : Ax\le b, \pi^T x \le \pi_0 \:\forall (\pi, \pi_0)\in\bigcup\nolimits_{i=1}^k S^i\}, \\
        &= \{x\in\mathbb{R}^n : A^k x\le b^k\}, \text{ with } A^0=A, b^0=b.\\
    \end{aligned}   
\end{equation}
The \emph{integrality gap} after separation round $k$ is given by the bound difference $g^k \coloneqq z^{\mathit{OPT}}-z^k\ge 0$.

To solve MILPs, cutting planes are often combined with a B\&B search procedure, in what is known as Branch-and-Cut (B\&C) algorithm: cuts are applied at the root node of the system (i.e., on the first LP relaxation of a MILP, as in \eqref{eq:relax}), and also in subsequent nodes to strengthen subproblems before branching. Given that tightening the relaxation before starting to branch is decisive to ensure an effective tree search, we focus on separation iteratively happening only at the root node of a B\&C system, i.e., on multiple cuts applied to \eqref{eq:relax}. 

Clearly, a pure cutting plane approach at the root node (without branching) is often not enough to optimally solve MILP instances, so only counting the number of added cuts does not give a complete picture of the cut selection performance. 
Instead, the \emph{integrality gap closed} (IGC) focuses on computing the factor by which the integrality gap is closed, between the first relaxation (solved on $P^0$) and the end of separation rounds  \citep[see also Section 2 of][]{TangAF20}. Formally, 
\begin{equation}\label{eq:igc}
    \text{IGC}^k\coloneqq\frac{g^0-g^k}{g^0}=\frac{z^k-z^0}{z^{\mathit{OPT}}-z^0}\in[0,1].
\end{equation}
When experimenting with a more general B\&C framework, performance measures typically focus on the total number of explored subproblems (nodes) and LP iterations needed to prove optimality. 

\paragraph{Cut Selection in SCIP}{
\looseness -1 State-of-the-art solvers like SCIP \cite{scip} can generate (or \emph{separate}) various families of cutting planes to populate a \emph{cut pool} \citep[see Section 3.3 of][]{Achterberg07}. 
In SCIP, various parameters are tuned to work in concert and control which types of cutting planes are separated and at what frequency, 
and also which ones are added to the formulation. Specifically, given a pool of available cuts, a parametrized scoring function ranks them, and a cut attaining maximum score is selected to be part of $S^k$. The rest of the cutpool is filtered for parallelism and quality;
remaining cuts are sorted again based on their initial score; and the next score-maximizing cut is selected, and so on, continuing until a pre-specified quota of cuts has been chosen for $S^k$, which is finally added to the LP relaxation.
At its core, then, cut selection proceeds by choosing \emph{one} cut at a time and is governed by a \emph{scoring function}, which in SCIP is a weighted sum of different scores from the MILP literature (integer support, objective parallelism, efficacy, directed cutoff)\footnote{See \url{https://scipopt.org/doc/html/cuts_8c_source.php\#l02509}.}, developed to empirically assess the potential quality of a given cut. We describe these scores and other cut selection heuristics in Appendix~\ref{app:baseline} (Table~\ref{tab:baselines}). 
}
\section{Learning to Cut by Looking Ahead}
\label{sec:main}

The central idea of our work is to learn a cut selection policy by imitation of a criterion that \emph{looks ahead} to explicitly measure the effect of a cut on the LP bound. In this section, after providing a precise formulation of the cut selection task, we describe and motivate our expert, and establish our imitation learning framework.

\paragraph{Core Cut Selection Task}{
Reviewing the cut selection process in SCIP, it becomes apparent that the core of cut selection resides in the ability to select a single cut from a set of available ones, via a scoring function. We focus precisely on this decision to formulate our cut selection task. Namely, given a pool of available cuts $\mathscr{C}$ for some relaxation $P$, the selection of a cut is made according to a \textit{scorer} criterion, which determines 
\begin{equation}\label{eq:argmax}
    C_{j^*} = \argmax_{C_j \in\mathscr{C}} {\mathit{scorer}(C_j, P)}.
\end{equation}
We consider multiple iterations of this basic decision, without introducing an additional notion of rounds: after a cut is selected as in \eqref{eq:argmax}, it is added to the formulation and the LP solved again, to produce a new pair $(\mathscr{C}', P')$ where $\mathscr{C}'$ is a freshly separated cutpool. In other words, we clear the previous $\mathscr{C}$ and replace it with cuts separated on $P'$.
Plugging-in different \textit{scorer} functions allows to obtain different cut selection criteria, some of which we describe in Appendix~\ref{app:baseline}.
}

\subsection{Designing a Lookahead Cut Selector}
\label{sec:LAdesign}

\looseness -1 As mentioned above, a natural way of assessing the impact of a cut is to measure the bound improvement it leads to in the relaxation---the bigger, the better. Of course, this metric cannot be evaluated without first trying out the cut itself, so MILP solvers make use of alternative indicators to gauge and approximate the cut effectiveness. 
We design a greedy criterion for cut selection performing such explicit \emph{lookahead steps} to measure beforehand the resulting LP bound of each cut available. Given $(\mathscr{C}, P)$, for a cut $C_j\in\mathscr{C}$ defined by $(\pi, \pi_0)$ we compare the relaxation given by $P$ with $P^{j}\coloneqq P\cap\{\pi^T x\le \pi_0\}$, obtained by adding $C_j$. In particular, we evaluate the dual bound difference of these two relaxations, which we call the \emph{lookahead (LA) score} of cut $C_j$:
\begin{equation}\label{eq:lascore}
    s_\textsc{la}(C_j, P)\coloneqq z^{j}-z\ge 0.
\end{equation}
Our \emph{Lookahead} cut selection criterion is obtained by using LA scores \eqref{eq:lascore} as the \textit{scorer} function in \eqref{eq:argmax}. 
Within a cutpool, it is common to observe multiple cuts with the same LA score, i.e., establishing the same bound improvement; in case of ties, \emph{Lookahead} breaks them randomly. 

The idea of performing a lookahead step is in the spirit of the strong branching (SB) heuristic \cite{ApplegateBCC07} for variable selection in B\&B: multiple works in the \emph{learning to branch} literature use this explicit evaluation as the costly but valuable expert to be imitated and ultimately replaced by a learned heuristic. 
Although improving the dual bound is a recognized goal in cut selection \cite{WesselmannS12,DeyM18}, to the best of our knowledge the explicit, SB-like approach has not been considered for cutting planes before.

\begin{figure}[t]
\vskip 0.2in
\begin{center}
\includegraphics{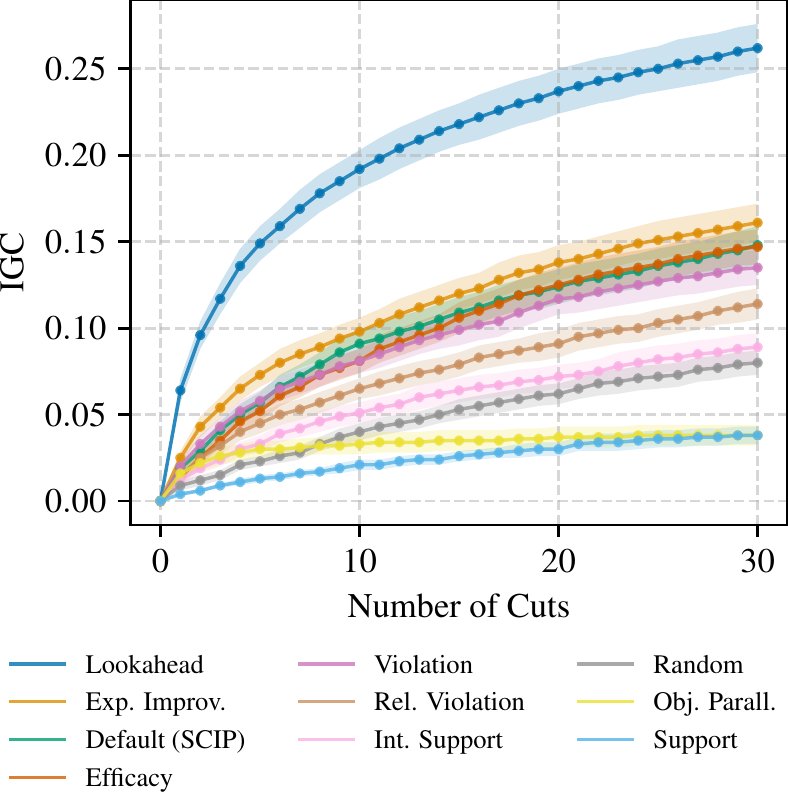}
\vskip -0.05in
\caption{Lookahead clearly outperforms common heuristics for cut selection on 510 instances from the MIPLIB (easy) collection. It achieves higher mean IGC throughout when performing 30 consecutive separation rounds and adding a single cut per round.}\label{fig:justification}
\end{center}
\vskip -0.2in
\end{figure}

\begin{figure*}[tb]
\vskip 0.2in
\begin{center}
\subfigure[\label{fig:encoding}]{\includegraphics[width=.52\textwidth]{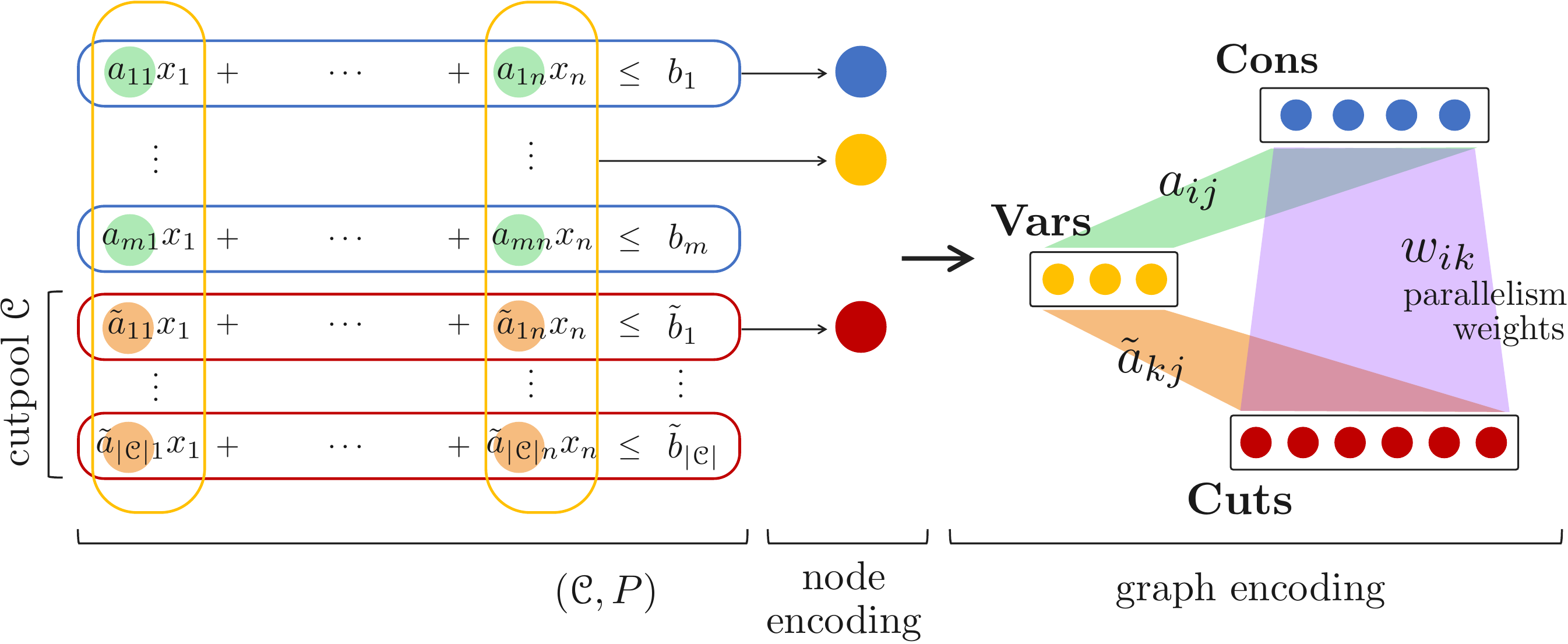}}
\hspace{5mm}
\subfigure[\label{fig:model}]{\includegraphics[width=.42\textwidth]{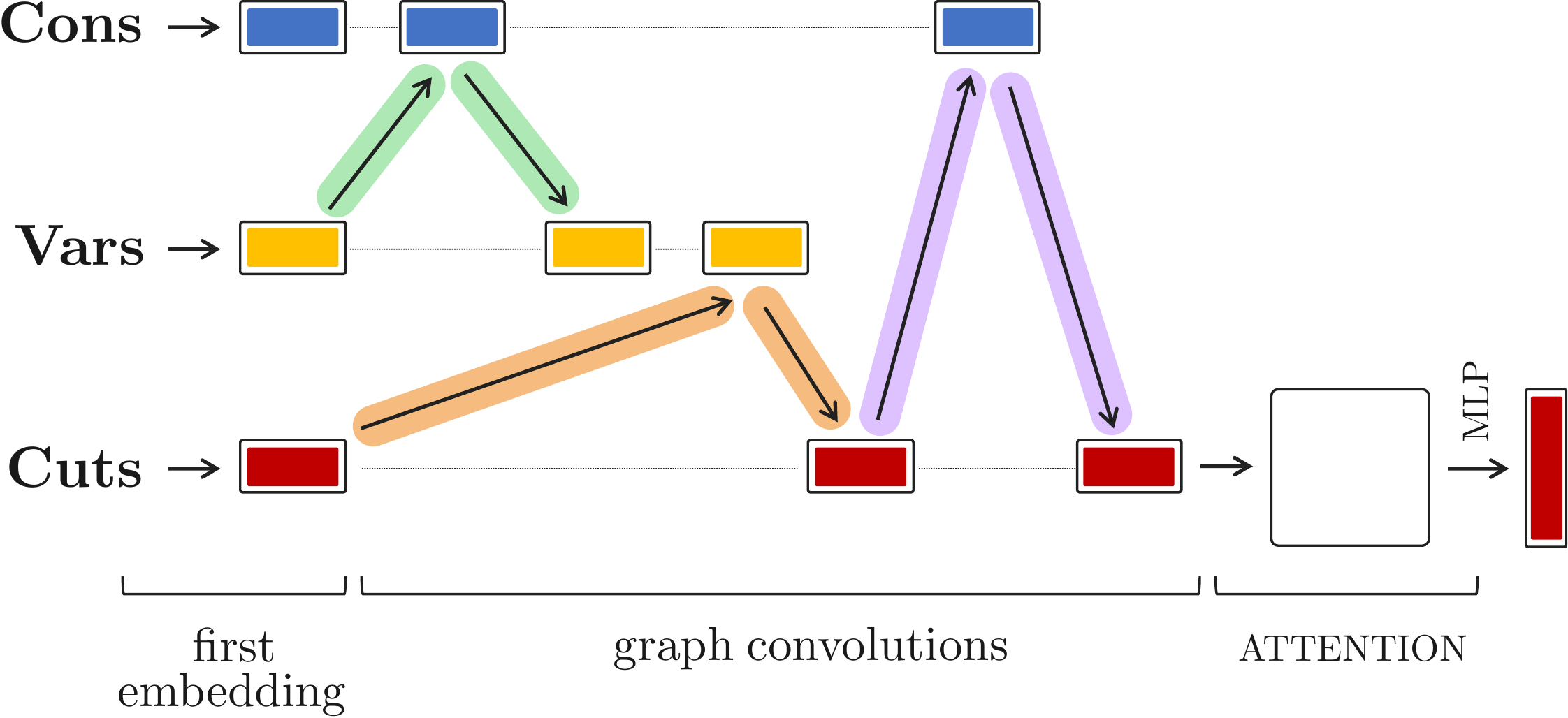}}
  \caption{(a) Node and graph encoding from the state $(\mathscr{C}, P)$. Parallelism weights are used on edges between constraints and cuts. (b) NeuralCut model: after three graph convolutions, an attention module allows cuts to attend to each other within the cutpool.}
  \label{fig:encoding-model}
\end{center}
\vskip -0.2in
\end{figure*}

\paragraph{A Strong (and Expensive) Criterion}{
We assess the proposed LA cut selection rule by exposing the basic cut selection task in a custom version of SCIP, and implementing scoring functions for common heuristics as well as SCIP's own \textit{scorer} formula (more specifications are given in Section~\ref{sec:experiments}). Operationally, to compute LA scores at a given relaxation $P$, we set SCIP in diving mode\footnote{Diving mode allows to temporarily alter some aspects of a problem (like enforcing a new constraint or variable bound), solve it, and be able to resume the original solving process.} to test one by one all candidate cuts in $\mathscr{C}$, and register their bound improvements.

We compare \emph{Lookahead} to common heuristics for cut selection on 510 bounded and feasible instances from the `easy' collection of MIPLIB 2017 \cite{gleixner2021miplib}. After pre-solving each instance, we perform 30 separation rounds and add a single cut per round to the LP relaxation. We average the IGC after each round over all instances and display the results in Figure~\ref{fig:justification}. Lookahead clearly outperforms other rules for cut selection throughout. It achieves larger IGC with fewer cuts than all baselines. The margin over the default rule of the SCIP solver is sizable; Lookahead achieves mean 0.25 IGC (compared to $\sim$0.15 IGC of the default rule) after 30 cuts and reaches 0.15 IGC after adding only five cuts on average. Further, since MIPLIB contains diverse instances that vary in size, complexity and structure, this suggest that \emph{Lookahead} may be a universally strong criterion for cut selection, potentially useful to improve performance in general-purpose MILP solvers.

\looseness -1 \emph{Lookahead} is thus a strong rule for selecting cuts---but an expensive one. At every iteration, running LA requires to solve $|\mathscr{C}|$ additional LPs; considering that even for small synthetic instances cutpools are typically populated by hundreds of cuts, repeating the LA iteration quickly becomes computationally intractable and makes the approach not viable in practice. 
In a way, \emph{Lookahead} appears to have all the credentials needed to become an \emph{expert} 
on which to train a ML policy.
}

\subsection{Lookahead Expert Imitation}
\label{sec:imitation}

In light of the cost of looking ahead, we propose to use LA scores to facilitate the training of a policy for cut selection via imitation learning. This learning approach offers a way to amortize the cost of expert decisions across entire families of MILP instances. 
By running several iterations of the basic cut selection loop, we can obtain a dataset of expert samples specified by $(\mathscr{C},P,\{s_\textsc{la}(C_j, P)\}_{C_j\in\mathscr{C}})$. We learn a cut selection policy that imitates the \emph{Lookahead} expert defined in Section~\ref{sec:LAdesign} by choosing the learnt scoring $\tilde{s}$ to minimize a soft binary entropy loss over all cuts $C \in \mathscr{C}$, 
\begin{equation}
 L\left(\tilde{s}\right) \coloneqq -\frac{1}{|\mathscr{C}|}\sum_{C \in \mathscr{C}} q_C\log\tilde{s}_C + \left(1 - q_C\right)\log(1-\tilde{s}_C)
\end{equation}
where $q_C = \frac{s_{LA}(C)}{s_{LA}(C^*_{LA})}$ and $C^*_{LA} = \arg\max_{C \in \mathscr{C}} s_{LA}(C)$.

\paragraph{State Encoding and Input Features}{
For each cut selection decision we extract information from the solver to describe the cutpool and LP relaxation pair $(\mathscr{C}, P)$ on which our \textit{scorer} operates. Specifically, we organize the state of the system as a tripartite graph whose nodes hold vectors (features) representing variables and constraints of $P$ (\textbf{Vars} and \textbf{Cons}), and cuts available in $\mathscr{C}$ (\textbf{Cuts}). Edges between \textbf{Vars} and \textbf{Cons} (resp. \textbf{Vars} and \textbf{Cuts}) are added when a variable appears in a constraint (resp. a cut), and carry the corresponding nonzero coefficients. We also link \textbf{Cons} and \textbf{Cuts} with a complete set of weighted edges. If $(a_i, b_i)$ is a \textbf{Cons} and $(\tilde{a}_k, \tilde{b}_{k})$ is a \textbf{Cut}, then the weight $w_{ik}\in[0,1]$ on the edge connecting them measures their degree of parallelism, i.e., $w_{ik} = |a_i^T\tilde{a}_k|/(\lVert a_i\rVert \lVert \tilde{a}_k \rVert)$.
In other words, hyperplanes are connected in the tripartite graph with a weight that measures 
their similarity. 
This facilitates a shortcut (skip connection) for communication between constraints and cuts, which could otherwise only happen via shared variables.
Figure~\ref{fig:encoding} illustrates node and graph encoding of the state $(\mathscr{C}, P)$. 

For input features, we extend those defined by \citet{GasseCFCL19} with new ones, to emphasize the role of constraints and cutting planes in our setup. For example, we add to our \textbf{Cons} and \textbf{Cuts} features the rank and fraction of nonzeros, as well as cut scores like violation, objective parallelism, support, and SCIP weighted score. A description of our input data is given in Appendix~\ref{app:features} (Table~\ref{tab:features}), where we also discuss feature implementation.
}

\paragraph{Policy}{
We use the graph encoding as input to NeuralCut, a new NN architecture for cut selection. The model we devise to parametrize a cut selection policy is a combination of a graph convolution neural network (GCNN) and an attention block on the cuts representations \cite{ScarselliEtAl09,VaswaniEtAl17}. 
Following the diagram in Figure~\ref{fig:model}, input features \textbf{Cons}, \textbf{Vars} and \textbf{Cuts} are first embedded into a hidden representation, before undergoing three passes of graph convolutions on the graph's edges. In order, the message passing happens between \textbf{Vars}$\rightarrow$\textbf{Cons}$\rightarrow$\textbf{Vars}; \textbf{Cuts}$\rightarrow$\textbf{Vars}$\rightarrow$\textbf{Cuts}; and finally \textbf{Cuts}$\rightarrow$\textbf{Cons}$\rightarrow$\textbf{Cuts}. Our rationale behind these convolution passes is to first get a sense of the current relaxation $P$ and cutpool $\mathscr{C}$, and then establish communication between available cuts and the constraints of the problem. 

A similar GCNN was proposed by \citet{GasseCFCL19} for B\&B variable selection: we change their model to introduce \textbf{Cuts} in our inputs and convolution passes, and replace their use of layer-norm with batch normalization \cite{BatchNorm}, which performed better in our setting. 

The last part of NeuralCut focuses on the cutpool representation, which enters an attention block to enable each cut to attend to all other cuts presently in the pool. We weigh the attention module with parallelism weights---calculated as the ones between a Cons and a Cut above---$w_{k,k'}$ between each pair of cuts $(k,k')$. A final MLP with sigmoid activation is used to predict a score $\tilde{s}\in[0,1]$ for each cut from its representation. 
Our implementation follows \citet{ShiEtAl21}, using PyTorch Geometric modules. 
}
\section{Experiments}
\label{sec:experiments}

We divide our computational experiments into two main parts. The first one focuses on evaluating learned NeuralCut policies on the core cut selection task: we measure the imitation learning performance of trained policies as well as their behavior when rolled-out on synthetic test MILP instances, in a controlled solver environment. Next, we investigates the impact that LA policies can have in a more realistic B\&C solver framework, with a dataset of real-world Neural Network Verification instances. 

\subsection{Evaluation Protocol}
\label{sec:protocol}

\paragraph{Solver Interface}{
\looseness -1 We derive a custom version of the SCIP solver (v.7.0.2) to expose and isolate the cut selection task as defined in Section~\ref{sec:main}. Our interface with the solver covers ten different separators, namely, \texttt{aggregation}, \texttt{clique}, \texttt{disjunctive}, \texttt{flowcover}, \texttt{gomory}, \texttt{impliedbounds}, \texttt{mcf}, \texttt{oddcycle}, \texttt{strongcg}, \texttt{zerohalf}. Heuristics for the \textit{scorer} function are implemented following \citet{WesselmannS12}, and we mirror the SCIP formula (SCIPScore). 
As previously mentioned, we exploit SCIP's diving mode to compute LA scores. The solver parametric setting is specified in Appendix~\ref{app:scip}.
}

\paragraph{Data Collection and Baselines}{
We experiment with different types of MILP models (see next sections). For each MILP dataset, we execute the first 10 iterations in our controlled cut selection environment to collect samples $(\mathscr{C},P,\{s_\textsc{la}(C_j, P)\}_{C_j\in\mathscr{C}})$. Ten cuts are sometimes enough to solve the easiest instances to optimality, and in general we observe that with more iterations the quality of cuts becomes more uniform within a cutpool.
To diversify the training space, during these 10 iterations we select cuts according to a \textit{scorer} uniformly drawn from $\{\text{Random, SCIPScore, \emph{Lookahead}}\}$. Note that regardless of the used \textit{scorer}, we always gather LA scores on the resulting $(\mathscr{C},P)$ state. 

The learned policy is compared to the \emph{Lookahead} expert and the SCIPScore baseline, as well as other \textit{scorer} heuristics including Efficacy, ExpImprovement, ObjParallelism, RelViolation, Violation, Support, IntSupport, Random. Again, we refer to Appendix~\ref{app:baseline} for details. 
}

\paragraph{ML Setup and Training}{
We train NeuralCut policies with ADAM \cite{ADAM} using the default PyTorch setting \cite{PyTorch}.
Our evaluations run on a distributed compute cluster; hardware specification are reported in Appendix~\ref{app:hardware}.
}

\subsection{Core Task Evaluation}
\label{sec:tangexp}

This first set of experiments is focused on assessing the quality of our imitation learning approach. We work in the controlled cut selection environment introduced above, and answer the following points: Are the proposed encoding and architecture (Figure~\ref{fig:encoding-model}) effective in modeling the task of cut selection, and do they enable the imitation of \emph{Lookahead}? Are trained NeuralCut policies successful in closing the integer gap when rolled-out on never seen test instances? We answer both questions positively on four different MILP benchmarks.

\paragraph{MILP Benchmarks}{
We consider the four classes of MILP models used in \citet{TangAF20} and reproduce their generators for instances of type Maximum Cut, Packing, Binary Packing and Planning; we refer to \citet{TangAF20} supplementary material for details on the models' mathematical formulations. A quick assessment revealed that instances of size Small and Medium were too easy for our setting, often being solved at presolve or after the addition of a handful of cuts. We thus focus on Large instances only, which exhibit sizes $n,m\in[50, 150]$, and generate a total of 3000 models for each family, divided into (2000, 500, 500) for (training, validation, test). Following the procedure outlined above, we gather just short of 20K samples for each training set, due to some models being solved in less than 10 cut selection iterations during data collection. 
Details about the composition of each benchmark are reported in Appendix~\ref{app:instances}. 
}

\paragraph{Test Evaluation Metrics}{
To measure the imitation performance of our policies, we design a \emph{bound fulfillment} metric assessing the quality of selected cuts. Specifically, for each collected test sample we look at its cutpool $\mathscr{C}$ and scale the attainable bound improvements within $[0,1]$, fixing at 1 the best one, specified by the cutpool's maximum LA score. Cut selection decisions for a trained policy and other heuristics can then be compared in this range to measure their ability in selecting bound-improving cuts. Note that \emph{Lookahead} achieves by definition a bound fulfillment of 1. 

We also plug-in the learned policies in our custom solver interface and evaluate their roll-outs on never-seen MILP test instances. For this online setting we employ IGC as defined in Section~\ref{sec:background} to track the integer gap, also measuring the area \emph{over} the IGC curve (the smaller, the better) to summarize each policy's performance. We call this metric \emph{reversed IGC integral}; note that in this case the minimum attainable integral is constrained by the composition of the cutpools. Runtime improvements are of marginal interest in diagnostic instances like those from \citet{TangAF20}, so we do not evaluate runtimes for this first set of experiments.
}

\begin{table*}[t]
\caption{NeuralCut selects cuts with high average bound fulfillment (left) and smaller average reversed IGC integral on four benchmarks. It outperforms both a competing RL approach \citep{TangAF20} and manual heuristics for cut selection. It approximates the performance of \emph{Lookahead} closely. Performance on test instances for 30 consecutive separation rounds and adding a single cut per round. Best model or heuristic is bold-faced, best overall is  in italics.}
\label{tab:core-all}
\begin{center}
\begin{small}
\adjustbox{max width=\textwidth}{%
\begin{tabular}{lcccccccccc}
\toprule
 && \multicolumn{4}{c}{\textbf{Bound fulfillment ($\uparrow$) on test samples}, mean} & & \multicolumn{4}{c}{\textbf{Reversed IGC integral ($\downarrow$) on test instances}, mean (ste)} \\
\noalign{\smallskip}\cline{3-6}\cline{8-11}\noalign{\smallskip}
 && \sc{\scriptsize Max. Cut} & \sc{\scriptsize Packing} & \sc{\scriptsize Bin. Packing} & \sc{\scriptsize Planning} & &  \sc{\scriptsize Max. Cut} & \sc{\scriptsize Packing} & \sc{\scriptsize Bin. Packing} & \sc{\scriptsize Planning}  \\
\midrule
\emph{Lookahead}       
&& \emph{1.0}   & \emph{1.0}   & \emph{1.0}   & \emph{1.0}
&& \emph{15.05 (0.09)} & \emph{26.42 (0.08)} & \emph{9.85 (0.33)} & \emph{10.33 (0.04)}\\
\midrule[0.2pt]
{NeuralCut}       
&& \textbf{0.96} & \textbf{0.61} & \textbf{0.78} & \emph{\textbf{1.0}} 
&& \textbf{15.55 (0.09)} & \emph{\textbf{26.30 (0.08)}} & \textbf{10.96 (0.33)} & \textbf{10.42 (0.04)}\\
{\citet{TangAF20}} 
&& 0.58 & 0.27 & 0.22 & 0.49
&&19.00 (0.09) & 27.59 (0.06) & 16.06 (0.38)  & 14.94 (0.08)  \\
\midrule[0.2pt]
{Default (SCIP) }      
&& 0.71 & 0.60 & 0.33 & 0.64     
&& 16.72 (0.09) & \emph{\textbf{26.29 (0.08)}} & 15.42 (0.28) & 14.01 (0.06)\\
{Exp. Improv.}  
&& 0.69 & 0.60 & 0.32 & 0.85     
&& 19.00 (0.08) & 26.27 (0.07) & 15.14 (0.29) & 11.15 (0.05)\\ 
{Efficacy }      
&& 0.65 & 0.60 & 0.32 & 0.46
&& 17.01 (0.09) & \emph{\textbf{26.28 (0.08)}} & 15.19 (0.29) & 14.52 (0.06)\\  
{Obj. Parall.}  
&& 0.47 & 0.34 & 0.27 & 0.44
&& 24.01 (0.08) & 28.28 (0.05) & 22.23 (0.26) & 27.71 (0.07)\\  
{Rel. Violation}   
&& 0.50 & 0.60 & 0.33 & 0.48     
&& 17.95 (0.08) & \emph{\textbf{26.28 (0.08)}} & 14.90 (0.29) & 15.36 (0.06)\\ 
{Violation}       
&& 0.64 & 0.35 & 0.21 & 0.26
&& 23.20 (0.08) & 28.81 (0.03) & 19.41 (0.31) & 25.75 (0.09) \\
{Support}        
&& 0.57 & 0.18 & 0.13 & 0.29     
&& 19.31 (0.10) & 28.77 (0.04) & 24.79 (0.22) & 18.39 (0.09)\\
{Int. Support}      
&& 0.62 & 0.18 & 0.21 & 0.34
&& 21.87 (0.07) & 28.78 (0.03) & 21.14 (0.28) & 23.31 (0.08)\\
{Random}          
&& 0.41 & 0.15 & 0.16 & 0.25     
&& 21.99 (0.07) & 28.73 (0.04) & 21.23 (0.28) & 23.26 (0.08)\\ 
\bottomrule	
\end{tabular}
} 
\end{small}
\end{center}
\vskip -0.1in
\end{table*}
\begin{figure}[t]
\vskip 0.2in
\begin{center}
  \centerline{\includegraphics{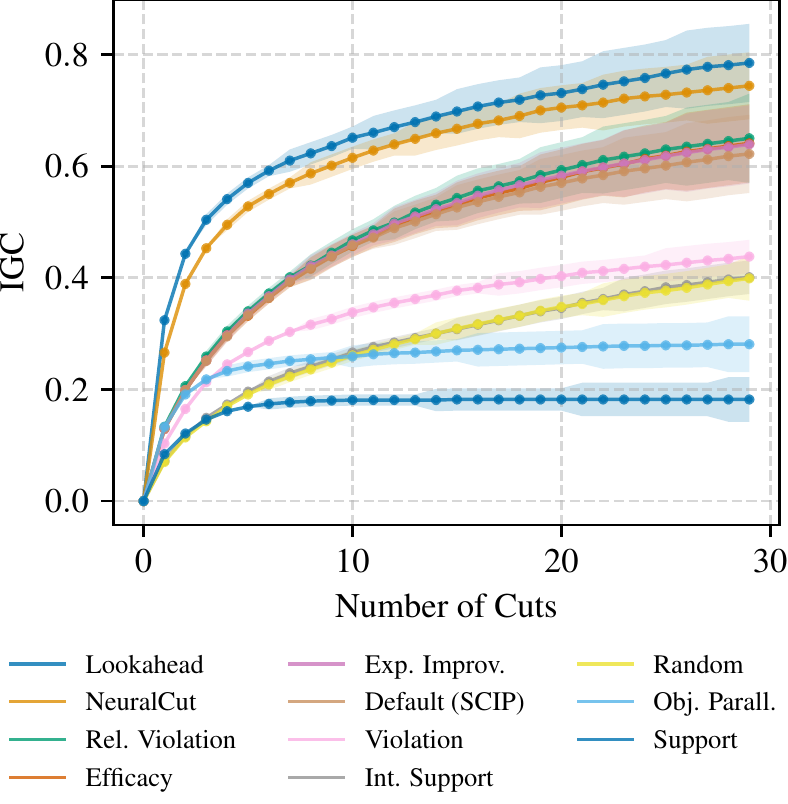}}
  \caption{On binary packing test instances, NeuralCut nearly achieves the same IGC as \emph{Lookahead} that it was trained to imitate. It outperforms other heuristics for cut selection. IGCs for the other benchmarks are in Figure \ref{fig:igc-all} in Appendix \ref{app:results:igc}.}
  \label{fig:tang-binpack-igc}
\end{center}
\vskip -0.2in
\end{figure}

\begin{table}[htb]
    \caption{Our model choices tend to improve tend to improve both bound fulfillment on test samples and reversed IGC integral on test instances for binary packing. Ablations for the other benchmarks are in Table \ref{tab:ablation-model-full} in Appendix \ref{app:results}.}
    \label{tab:binpack-ablation}
    \begin{center}
    \begin{small}
    \adjustbox{max width=\textwidth}{%
    \begin{tabular}{lcc}
        \toprule
         & \textbf{Bound full.} &  \textbf{Rev. IGC integral} \\
        \midrule
        {\citet{GasseCFCL19} }     & 0.54 & 16.09 (0.31) \\
        { $+ \text{ our features}$}   & 0.66 & 12.77 (0.32) \\
        { {$+ \text{ our architecture}$}}    & 0.76 & \textbf{10.73 (0.33)} \\
        \midrule[0.2pt]
        {NeuralCut ($+ \text{ our graph}$)} & \textbf{0.78} & \textbf{10.96 (0.33)} \\
        \bottomrule
    \end{tabular}
    } 
    \end{small}
\end{center}
\vskip -0.1in
\end{table}

\paragraph{Results}{
Table~\ref{tab:core-all} reports results on the evaluation of NeuralCut policies trained on each MILP family. 
Bound fulfillment rates (left side of the table) confirm that NeuralCut effectively learns to imitate \emph{Lookahead}, selecting bound-improving cutting planes better than all other heuristic baselines including SCIPScore, and scoring almost perfectly on Max. Cut and Plan datasets.
Even though the difficulty of the selection task may vary across benchmarks, as reflected by the average size of the cutpools in the test samples, the imitation learning performance is quite impressive and homogeneous. 
In terms of integer gap closed over 30 iterations roll-outs (i.e., on the addition of the first 30 cuts) (right side), on 3/4 datasets NeuralCut achieves smallest reversed IGC integrals, the nearest to \emph{Lookahead}; overall, we observe a good correlation with the bound fulfillment metric. The Packing case is tied between many rules, and does not seem to offer much space for improvement via LA greedy selection. 
Figure~\ref{fig:tang-binpack-igc} depicts mean IGC curves for NeuralCut on binary packing instances, and clearly shows the advantage of the LA-trained policy over all baselines; similar plots for the other benchmarks are collected in Appendix~\ref{app:results}. 

We also compare NeuralCut to the reinforcement learning (RL) approach for cut selection proposed by \citet{TangAF20}. They cast cut selection as an RL problem and use improvements to the LP bound as rewards to train a model for cut selection via evolutionary strategies. \citet{TangAF20} consider only Gomory cutting planes and implement their model with Gurobi \cite{gurobi}. To compare to them, we re-implement their method in our SCIP solver interface and expose it to all cutting planes. NeuralCut outperforms RL on all four benchmarks (Table~\ref{tab:core-all}). It achieves higher bound fulfillment and lower reversed IGC integral. The difference is particularly acute on the binary packing instances (0.78 \emph{vs} 0.22 bound fulfillment and 19.96 \emph{vs} 10.96 \emph{vs}  16.06 reversed IGC integral).}. 

With respect to ML modelling choices previously used for MILP optimization, NeuralCut introduces several innovations. These include new features, the introduction of short-cut connections between constraints and cuts and the use of an attention module for the cutpool. In Table \ref{tab:binpack-ablation}, we perform model ablations on the binary packing instances, to better assess the effectiveness of our modelling choices. We begin with a simple GNN model that is based on \citet{GasseCFCL19}, but adapted to cut selection by using cuts in place of constraints in the original model formulation of \citet{GasseCFCL19} and switching the order in which half-convolutions are applied. Second, we use the features of NeuralCut ($+ \text{ our features}$) for both cuts and variables instead of the features of \citet{GasseCFCL19}. Third, we additionally make architectural choices, i.e. we replace layer normalization with batch normalization and add the attention module to the model. Finally, we leverage the tri-partite graph formulation to model constraints explicitly (in addition to the cuts and variables), which gives the full NeuralCut model. For binary packing, improvements are most pronounced when using our features (0.54 \emph{vs} 0.66, 16.09 \emph{vs} 12.77) and applying our architectural choices (0.66 \emph{vs} 0.76, 12.77 \emph{vs} 10.73). Ablations for the other benchmarks are in Table \ref{tab:ablation-model-full} in Appendix \ref{app:results:modelablation}. In addition, an ablation on the choice of loss function is in Table \ref{tab:ablation-full-loss} in Appendix \ref{app:results:lossablation}.

\begin{table*}[tb]
    \caption{NeuralCut in a B\&C solver -- Median metrics for NN verification test instances and different values for the threshold parameter $\epsilon$. NeuralCut improves the dualbound at the root at a fraction of the number of cuts the default solver SCIP selects. When subsequently branching from the root node, this can reduce the remaining solving time by lowering the number of LP iterations or expanded nodes in the search tree. Best are bold-faced.}
    \label{tab:nnverification}
    \vskip 0.15in
    \begin{center}
    \begin{small}
    \begin{tabular}{lccccc}
        \toprule
                & \textbf{\# cuts} & \textbf{Rel. bound improv.} & \textbf{Time (s)} & \textbf{\# nodes} & \textbf{\# LP iters.}\\ 
        \midrule
        SCIP B\&C   & 279 & \textbf{1.00} & 23.65 & 745 & 21933\\
        \midrule 
        NeuralCut, $\epsilon = 10^{-5}$     & 105 & \textbf{1.00} & 22.35 & \textbf{671} & 18846\\
        NeuralCut, $\epsilon = 10^{-4}$     & 81 & 0.99 & \textbf{20.89} & 756 & 19310\\
        NeuralCut, $\epsilon = 10^{-3}$     & 48 & 0.98 & 22.73 & 803 & \textbf{18048}\\
        NeuralCut, $\epsilon = 10^{-2}$     & 27 & 0.94 & 24.06 & 861 & \textbf{18048}\\
        NeuralCut, $\epsilon = 10^{-1}$     & 11 & 0.76 & 25.09 & 998 & 21996\\
        NeuralCut, $\epsilon = 1$     & \textbf{10} & 0.53 & 23.35 & 952 & 21611\\
        \bottomrule
    \end{tabular}
    \end{small}
\end{center}
\vskip -0.1in
\end{table*}

\subsection{NeuralCut in a B\&C Solver}
\label{sec:spotlight}

Up to this point we worked in a controlled environment and focused on the core task of selecting a single cut from a pool of available ones. While for the sake of evaluation clarity we isolated this decision from all the other choices related to cuts management happening in a solver, in practice these heuristics work in concert and impact each other. 

After having determined the effectiveness of NeuralCut policies in the controlled environment, we now wish to investigate the impact that selecting good (bound-improving) cuts can have on a more complex and realistic solver setting. To do so, we deploy NeuralCut policies in the SCIP original B\&C framework, and compare them against the default solver on a challenging MILP dataset of neural network verification instances. 

\paragraph{NN Verification Dataset}{
The problem of verifying the robustness of a neural network to input perturbation can be formulated as a MILP \cite{ChengEtAl17,TjengEtAl18}. Because each input to be verified gives rise to a different instance, this application is particularly suited to get large MILP benchmarks that are not synthetic. The dataset we use to stress-test our policies 
was open-sourced by \citet{NairEtAl20}.\footnote{\url{https://github.com/deepmind/deepmind-research/tree/master/neural_mip_solving}}
Compared to synthetic instances, these MILPs are larger: \citet{NairEtAl20} report median sizes $(n,m)$ at (7142, 6531) and harder to solve for SCIP.
Interestingly for us, the formulation of these models is notoriously weak, i.e., a lot of work needs to be done to tighten the dual bound of the LP relaxation.
From the original dataset splits we 
select 1807 problems (Appendix~\ref{app:instances}).
}

\paragraph{Solver Plug-in}{
To deploy our policies in SCIP we implement a plug-in that gets called during separation rounds;\footnote{Our plug-in is inspired by \url{https://github.com/avrech/learning2cut}.} we minimally interfere with the solver's default process and only operate at the root node level. In particular, we do \emph{not} override SCIP separation and cuts management parameters, and only force the sorting of SCIP cutpools based on our predictions.
In this realistic solver environment, our models also need to make the choice of whether to add a cut: we combine NeuralCut with a simple stalling rule parametrized on a threshold parameter $\epsilon$, stopping the selection after the bound improvement from adding cuts has not exceeded $\epsilon$ for 10 consecutive rounds.
Technical details, solver parametric setting and changes to data collection are discussed in Appendix~\ref{app:scip}.
}

\paragraph{Test Evaluation Metrics}{

We assess NeuralCut policies' performance in the B\&C system with (1) measures collected at the end of the root node, like total number of cuts and bound improvement relative to SCIP (1.0 = same bound as SCIP), and (2) end-of-run statistics, like solving time, number of nodes and number of LP iterations, which we collect when the MILP is subsequently solved. 
Wanting to highlight the potential of improving B\&C solvers with ML models for cutting planes selection, 
for the second set of metrics we factor out the root node contributions, where NeuralCut operates. This allows us to get a clearer picture of the impact that an improved cut selection at the root node can have on the remaining solving process. 
Because of the presence of outliers in the dataset, we report median statistics.
}

\begin{figure}[tb]
\begin{center}
  \centerline{\includegraphics{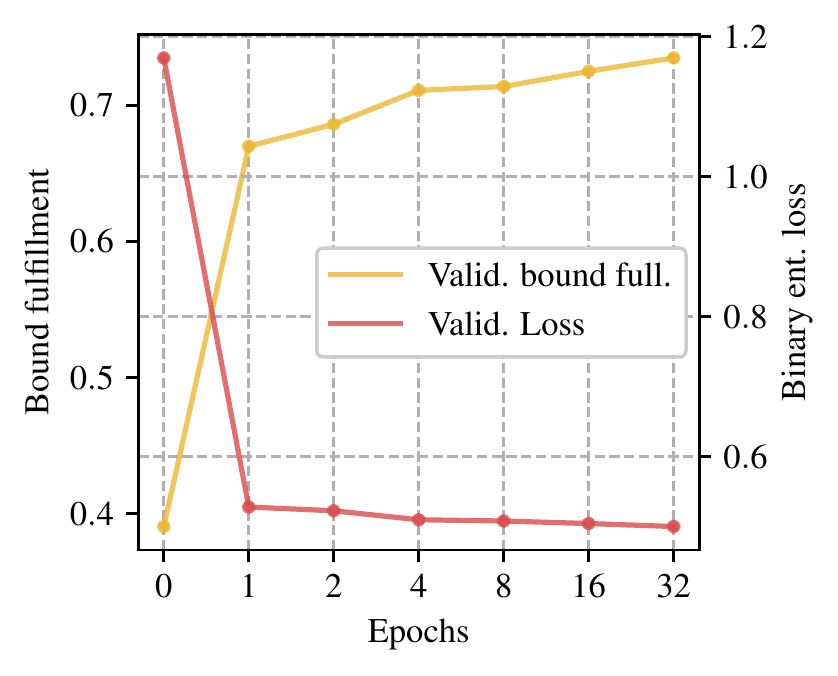}}
  \caption{NeuralCut attains good validation performance after training only for one epoch. Validation performance improves as training continues.}
  \label{fig:learningcurves}
\end{center}
\end{figure}

\paragraph{Results}{
Figure~\ref{fig:learningcurves} shows learning curves (binary entropy loss and bound fulfillment) for a NeuralCut policy trained on NN verification for 32 epochs. The model learns to select good cuts: it achieves bound fulfillment score of over 70\% from an initial 38\%, confirming the ability of NeuralCut to scale to real-world instances.
Table~\ref{tab:nnverification} summarizes the performance of NeuralCut policies compared to the default B\&C solver. 
At the root node, NeuralCut reaches the same or very similar median bound while only adding a fraction of SCIP's cut budget (20-40\%). This cut efficiency at the root node translates into improved solving behaviour after the root node relaxation. 
When subsequently solving the MILP, NeuralCut's additions help to improve the median remaining solving time by up to 10\% (in NeuralCut with $\epsilon=10^{-4}$), by reducing the number of nodes in the search tree and the number of LP iterations. 
Overall, results on the realistic solver setting validate the positive effects of selecting bound-improving cuts which were already established in Section~\ref{sec:tangexp}, and exhibit the potential of ML methods in this complex setting.

}
\section{Related Work}
\label{sec:related}
Recent years have seen a rise in the application of ML approaches to the MILP algorithmic framework \cite{BengioLP21}, and imitation learning has been identified as a natural approach for tasks in the field \cite{MAlvarezLW17,HansknechtJS18,GasseCFCL19,Zarpellon_Jo_Lodi_Bengio_2021}. 
On cutting planes, \citet{RaduBMT18} first introduced a NN estimator predicting the objective improvement of a cut in the context of semidefinite programming relaxations and quadratic optimization. Closer to our setting, the work of \citet{TangAF20} proposes a reinforcement learning framework for the Gomory cutting plane algorithm, where policies are trained via evolutionary strategy. \citet{HuangEtAl21} instead frame cut selection as multiple instance learning, and apply it to a proprietary solver. On the theoretical side, \citet{BalcanPSV21} produce provable guarantees for learning cut selection policies. 

The idea of performing explicit lookahead steps was introduced in the MILP literature with Strong Branching \cite{ApplegateBCC07}, and subsequently explored by \citet{GlankwamdeeL11} and \citet{Schubert17}, always in the context of variable selection. Related to explicitly optimizing cutting planes on their bound improvements, relevant works are \cite{AmaldiCG14} and \cite{ConiglioT15}, the latter proposing a ``bound optimal'' cutting plane \emph{separation} algorithm. 

Finally, GNNs have been used before to approach combinatorial tasks \cite{CappartEtAl21}, and part of our model is similar to the one used by \citet{GasseCFCL19} for learning to branch. 
Attention mechanisms and Transformer architectures were first proposed in the Natural Language Processing field \cite{VaswaniEtAl17}, and are being rapidly adopted in several other applications.

\section{Conclusions}
\label{sec:conclusions}
The task of cutting planes selection is essential for solving MILPs and we presented a new imitation learning framework to tackle it. 
We defined a \emph{Lookahead} criterion that greedily selects cuts based on LP bound gains: we showed that this rule delivers strong decisions for cut selection and appears more effective than common heuristics, but with the drawback of being too expensive to be viable in practice. Calling ML to the rescue, we used LA scores to train a policy via imitation learning. We organized the cut selection state in a graph structure and developed a novel NN architecture to predict scores over a cutpool. Controlled experiments on four synthetic families of MILPs show how our NeuralCut policies outperform standard heuristics, proving effective in imitation and successful in closing the integrality gap when rolled-out on never-seen test instances. A stress-test on the challenging benchmark of NN Verification models further validates our approach, and demonstrates the positive effects of learning to select good (bound-improving) cuts at the root node of a realistic B\&C framework. Our results highlight the potential for improving solver performance with learned models for cut selection.

The heuristic nature and complexity of MILP cuts management leaves plenty of space for ML approaches. Interesting questions include the decision of whether to trigger separation and when to stop cut addition. Exploring cutting planes at local nodes could instead allow to study the generalization ability of policies trained at the root of a B\&C tree. Finally, because \emph{Lookahead} appears to be a universal criterion for cut selection, it could potentially be used to improve performance in the general-purpose setting, too. We believe that investigating strategies to achieve broader generalization of ML for MIP (e.g., via diverse training sets, surrogate losses, model architectures) is an interesting agenda for future research.

\section*{Acknowledgements}
We would like to thank Antoine Prouvost, Andrea Lodi and Avrech Ben-David for helpful discussions on the topic. MBP gratefully acknowledges support from the Max Planck ETH Center for Learning Systems. Resources used in preparing this research were provided, in part, by the Sustainable Chemical Processes through Catalysis (Suchcat) National Center of Comptence in Research (NCCR), the Province of Ontario, the Government of Canada through CIFAR, in particular the CIFAR AI Chairs and CIFAR AI Catalyst Grants programs, and companies sponsoring the Vector Institute. Finally, we thank Sam Fux from the High Performance Computing group at ETH Z\"urich for his help with getting things running on the ETH Euler cluster. 
\bibliography{tex/refs/refs}
\bibliographystyle{tex/sty/icml2022}
\newpage
\appendix
\onecolumn
\section{MILP Instances}
\label{app:instances}
\subsection{MIPLIB 2017}
We used 510 instances of the MIPLIB 2017 collection \citep{gleixner2021miplib}. We only considered instances that were considered `easy' by \citep{gleixner2021miplib} in spring 2022, bounded and feasible, such that IGC could be readily computed. For 119 out of 629 `easy', bounded and feasible instances, we were unable to perform the total 30 separation rounds (with all cutting plane generators) on standard compute hardware within 24 hours and excluded these instances from consideration. 

\subsection{Benchmarks}
We used instances from the four integer programming domains maximum cut, packing, binary packing and planning as suggested by \citet{TangAF20} for cut selection. Their exact mathematical formulation is given in \citet{TangAF20}. They consider three different sizes (\emph{small}, \emph{medium}, \emph{large}) for each domain. Early experiments revealed that \emph{small} and \emph{medium}-sized instances were often pre-solved or easily solvable by adding a very small number of cuts. Therefore, we only considered \emph{large} instances. For each domain, we generated 1000 instances for training, 500 instances for validation and 500 instances for testing. To generate these instances, we followed the protocol in \citep{TangAF20}. We used graphs with $|V|=14$ vertices and $|E|=40$ edges for maximum cut, packing problems with $m=60$ (resource) constraints $n=60$ variables, binary packing problems with $m=66$ (resource) constraints, $66$ binary constraints and $n=66$ variables and a time horizon $T=40$ for production planning.

\subsection{Neural Network Verification}
We also work with a neural network verification dataset from \citet{NairEtAl20}, which is obtained from the verification of a trained convolutional NN on each image in the MNIST dataset. The corresponding MILP mathematical formulation is given by \citet{GowalEtAl19}. We select from the original MILP collection a total of 1807 problems: we exclude infeasible instances (often trivially solved at presolve) and those models hitting a 1 hour time-limit in SCIP default mode. We considered 1260 instances for training, 271 instances for validation and 276 instances for testing.

\section{Baseline Heuristics}\label{app:baseline}

A description of baselines heuristics used for cut selection and their SCIP implementation details are given in Table~\ref{tab:baselines}. In particular, note that while our \emph{Lookahead} rule computes the bound improvement exactly, the \emph{expected improvement} heuristic (ExpImprovement) aims at estimating the LP gain of a cut (under certain assumptions), and is derived as a product of other two scores. 
A useful reference is \cite{WesselmannS12}. 

\begin{table}[tbh]
    \caption{Description of baselines heuristics used for cut selection.}
    \label{tab:baselines}
    \vskip 0.15in
    \begin{center}
    \begin{small}
    \begin{tabularx}{0.9\textwidth}{lp{0.7\textwidth}}
    \toprule
        \textbf{Name} & \textbf{Description} \\
    \midrule
        Random & Random cut selection.\\
        \midrule
        Violation & Infeasibility of a cut with respect to the current LP solution. \newline Computed as $-$SCIProwGetLPFeasibility.\\
        \midrule
        RelViolation & Relative violation, i.e., violation score scaled by $\min$ of a cut rhs ($\pi_0$) and lhs ($\pi^Tx$).\\
        \midrule
        Efficacy & Cut's efficacy with respect to the current LP solution. \newline Computed as  $-$SCIProwGetLPFeasibility scaled by the cut's euclidean norm.\\
        \midrule
        ObjParallelism & Parallelism of a cut with the objective function. \newline Computed as SCIProwGetObjParallelism.\\
        \midrule
        ExpImprovement & Expected improvement, i.e., squared Euclidean norm of objective function vector multiplied by objective parallelism and efficacy. \newline Computed as objsqrnorm $\cdot$ SCIPgetCutObjParallelism $\cdot$ SCIPgetCutEfficacy.\\
        \midrule
        Support & Support score, negative proportion of a cut nonzero support. \newline Computed as $-$SCIProwGetNNonz scaled by the number of variables.\\
        \midrule
        IntSupport & Integer support, i.e., proportion of cut's integer entries over nonzero ones. \newline Computed as SCIProwGetNumIntCols / SCIProwGetNNonz. \\
        \midrule
        SCIPScore & SCIP default score, i.e., weighted sum of obj. parallelism, integer support and efficacy scores.\\
    \bottomrule
    \end{tabularx}
    \end{small}
    \end{center}
    \vskip -0.1in
\end{table}

\section{Input Features}\label{app:features}

We interface SCIP and collect input features via custom PySCIPOpt functions, closely following the structure of Ecole's NodeBipartite observation function\footnote{\url{https://github.com/ds4dm/ecole/blob/master/libecole/src/observation/node-bipartite.cpp}} \cite{ProuvostEtAl20}. With respect to the features defined in \cite{GasseCFCL19}, we remove from \textbf{Vars} two attributes related to incumbent values (incval and avgincval), as very rarely an incumbent has been found already at the root node for our problems and parametric setting. We substantially extend the parametrization of constraints and cutting planes. Note that we represent both types of rows by the same feature set, even though parts of the feature (e.g., scores) are more meaningful for just one of the two types (e.g., cuts). Our input state for a cut selection sample on $(\mathscr{C}, P)$ is thus composed of $\mathbf{Vars}\in\mathbb{R}^{n\times17}$, $\mathbf{Cons}\in\mathbb{R}^{m\times34}$ (where $n,m$ are the number of variables and constraints in $P$), and $\mathbf{Cuts}\in\mathbb{R}^{|\mathscr{C}|\times34}$. We attach to edges of the graph encoding scalar row coefficients and, for each pair of (cut, constraint) their parallelism coefficient as weight. A detailed description of the input features can be found in Table~\ref{tab:features}.

\begin{table}[th]
    \caption{Description of features populating the input's vectors.}
    \label{tab:features}
    \vskip 0.15in
    \begin{center}
    \begin{small}
    \begin{tabular}{cll}
        \toprule
        \textbf{feature} & \textbf{Feature} & \textbf{Description} \\
        \midrule
        \multirow{11}{*}{\textbf{Vars}} & norm\_coef & Objective coefficient, normalized by objective norm\\
        & type & Type (binary, integer, impl. integer, continuous) one-hot\\
        & has\_lb & Lower bound indicator\\
        & has\_ub & Upper bound indicator\\
        & norm\_redcost & Reduced cost, normalized by objective norm\\
        & solval & Solution value\\
        & solfrac & Solution value fractionality\\
        & sol\_is\_at\_lb & Solution value equals lower bound\\
        & sol\_is\_at\_ub & Solution value equals upper bound\\
        & norm\_age & LP age, normalized by total number of solved LPs\\
        & basestat & Simplex basis status (lower, basic, upper, zero) one-hot\\
        \midrule
        \multirow{24}{4em}{\textbf{Cons}, \textbf{Cuts}} & is\_cut & Indicator to differentiate cut vs. constraint\\
        & type & Separator type, one-hot\\
        & rank & Rank of a row\\
        & norm\_nnzrs & Fraction of nonzero entries\\
        & bias & Unshifted side normalized by row norm\\
        & row\_is\_at\_lhs & Row value equals left hand side\\
        & row\_is\_at\_rhs & Row value equals right hand side\\
        & dualsol & Dual LP solution of a row, normalized by row and objective norm\\
        & basestat & Basis status of a row in the LP solution, one-hot\\
        & norm\_age & Age of row, normalized by total number of solved LPs\\
        & norm\_nlp\_creation & LPs since the row has been created, normalized\\
        & norm\_intcols & Fraction of integral columns in the row\\
        & is\_integral & Activity of the row is always integral in a feasible solution\\
        & is\_removable & Row is removable from the LP\\
        & is\_in\_lp & Row is member of current LP\\
        & violation & Violation score of a row\\
        & rel\_violation & Relative violation score of a row\\
        & obj\_par & Objective parallelism score of a row\\
        & exp\_improv & Expected improvement score of a row\\
        & supp\_score & Support score of a row\\
        & int\_support & Integral support score of a row\\
        & scip\_score & SCIP score of a row for cut selection\\
        \bottomrule
    \end{tabular}
    \end{small}
    \end{center}
    \vskip -0.1in
\end{table}




\section{SCIP Interface and Parametric Settings}\label{app:scip}

\paragraph{Core Task Evaluation (Section~\ref{sec:tangexp})}{
Our controlled SCIP environment covers separators: \texttt{aggregation}, \texttt{clique}, \texttt{disjunctive}, \texttt{flowcover}, \texttt{gomory}, \texttt{impliedbounds}, \texttt{mcf}, \texttt{oddcycle} (not enabled in SCIP default), \texttt{strongcg}, \texttt{zerohalf}. It runs with presolve enabled and without primal heuristics; we override SCIP's separation and enforce instead our cut selection loop on the 10 custom separators, by setting the node limit to 1, disabling SB computations and fixing \texttt{minefficacy} parameters to 0, to include all separated cuts in the pool. Because in our experiments we disable all primal heuristics, we mirror SCIP's formula (SCIPScore) without including the directed cutoff metric, i.e., as it is computed in the solver when no primal solution is available.
}

\paragraph{NeuralCut in a B\&C Solver (Section~\ref{sec:spotlight})}{
The plug-in for experiments in SCIP minimally interferes with the default optimization process. With respect to the SCIP procedure described in Section~\ref{sec:background}, we only get input features, LA scores (when collecting data), and force the sorting of SCIP cutpools based on our predictions, so that SCIP automatically selects the cut identified by LA/a trained policy. We only operate at the root node separation level. We parametrize the stalling rule to operate NeuralCut policies with two parameters: a tolerance $\epsilon$ to define stalling on bound improvement, and the number $K$ of consecutive stalling rounds. In our experiments, we fix $K=10$, similarly to SCIP's own parameter.

For data collection in this setting, we use SCIP default rule as the ``exploration agent'' and collect data within SCIP rounds with 1\% probability; we fix a low collection rate to keep our dataset small and diverse, as we found there is little information gained from storing samples from highly correlated rounds.
In terms of parametric setting, we run experiments with presolve and primal heuristics enabled, no node limit, and a time-limit of 1h; we disable restarts and reoptimization. In particular, we do \emph{not} override SCIP separation and cuts management parameters. 
}

\section{Hardware Specifics}\label{app:hardware}
We collect data for training and evaluate all methods on a distributed compute cluster which predominantly contains Intel Xeon E3-1585Lv5 CPUs. A single GPU device (NVidia GeForce GTX 1080 Ti) is only used for training the models, but not for evaluation.

\section{Additional Results}
\label{app:results}
We include additional experimental results in this section. We visualize the IGC trajectories of Lookahead, NeuralCut and manual heuristics on the four benchmarks in \ref{app:results:igc}. We test the transferability of NeuralCut models across different domains of integer programs in \ref{app:results:transfer}. We perform model ablations in \ref{app:results:modelablation}. We perform ablations on the loss function of the surrogate in \ref{app:results:lossablation}.
\subsection{Integrality Gap Closed}
\label{app:results:igc}
\begin{figure}[h]
\label{fig:igc-all}
  \caption{Mean test IGC curves for NeuralCut models trained on Maximum Cut, Packing, Binary Packing and Planning instances.}
\vskip 0.2in
\begin{center}
\subfigure[\label{fig:igc-maxcut}\normalsize{Maximum Cut}]{\includegraphics{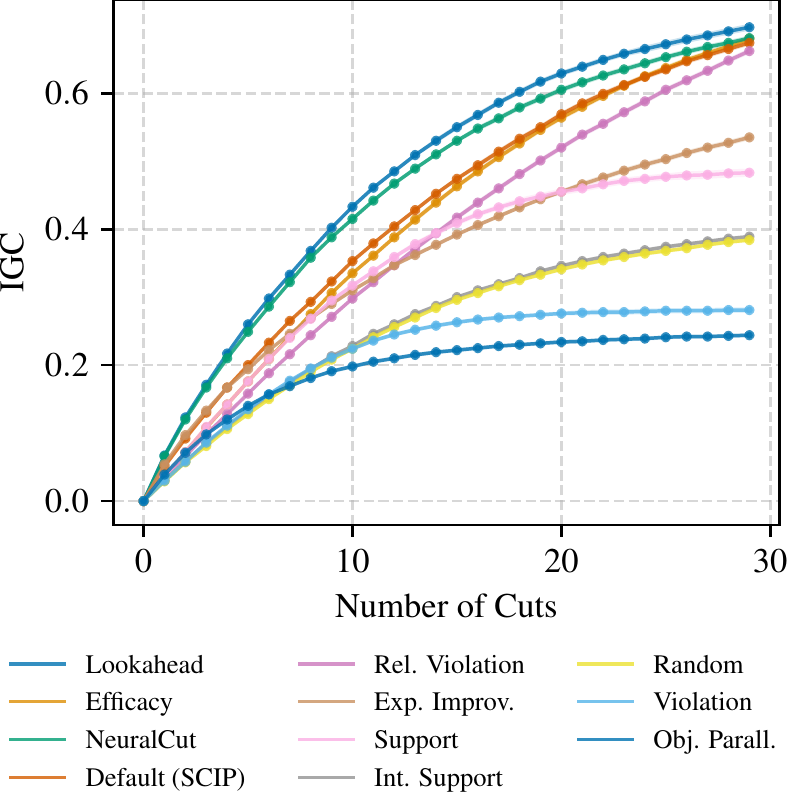}}
\hfill
\subfigure[\label{fig:igc-packing}\normalsize{Packing}]{\includegraphics{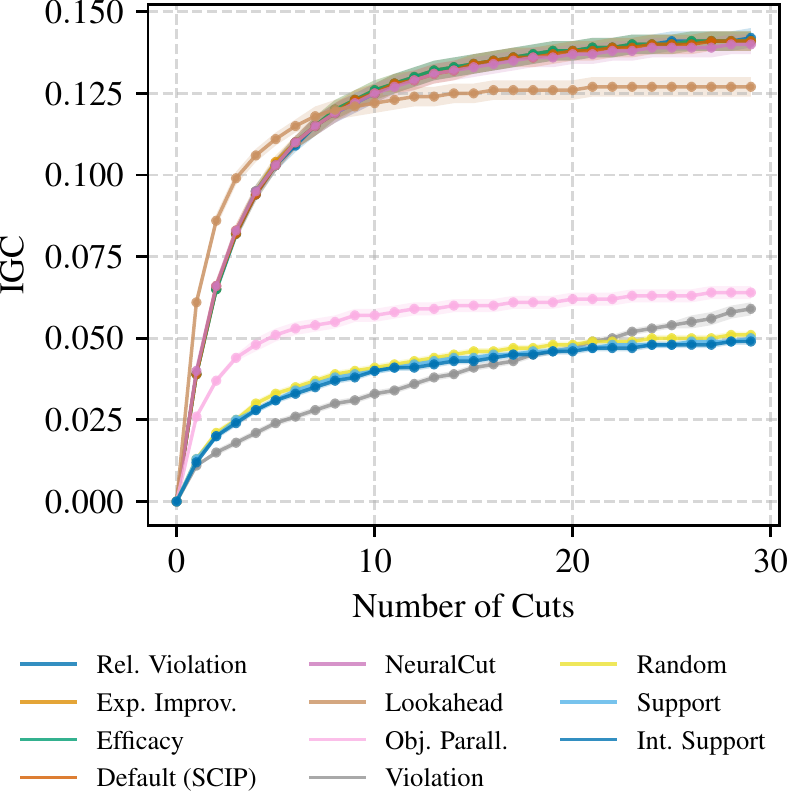}}
\subfigure[\label{fig:igc-binpacking}\normalsize{Binary Packing}]{\includegraphics{include/graphics/igc-binpacking.pdf}}
\hfill
\subfigure[\label{fig:igc-planning}\normalsize{Planning}]{\includegraphics{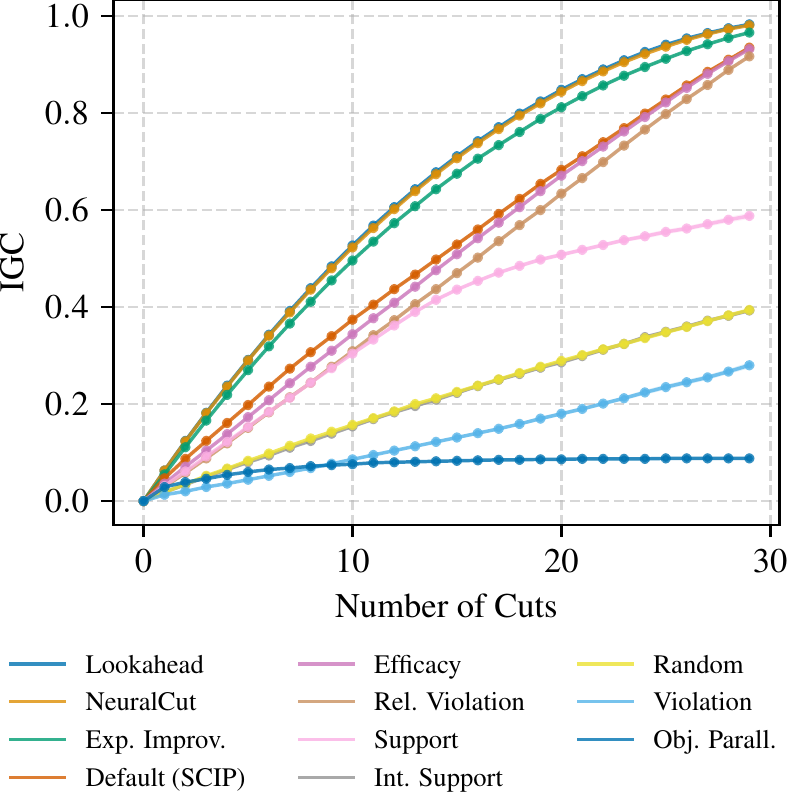}}
\end{center}
\vskip -0.2in
\end{figure}

In Figure \ref{fig:igc-all}, we visualize the IGC trajectories of Lookahead, NeuralCut and common heuristics for cut selection on the four benchmarks Maximum Cut, Packing, Binary Packing and Planning. The NeuralCut models correspond to those in Table \ref{tab:core-all}. All models and heuristics were evaluated on the 500 test instances of the respective domain. 30 separation rounds were performed in each a single cut was added.

\subsection{Transferability}
\label{app:results:transfer}
In Table \ref{tab:transfer}, we test the transferability of NeuralCut models across different domains of integer programs. For this purpose, each of the four NeuralCut models in Table \ref{tab:core-all} is additionally evaluated on the test instances of the other three domains. For example, the second-row (packing) fourth-column entry of Table \ref{tab:transfer} corresponds to the NeuralCut model trained on packing instances, but evaluated on planning test instances. We find that NeuralCut models do not transfer well to other domains: On all benchmarks, the test performance of transferred models is significantly worse than the performance of the model that was trained on instances of the respective domain. This suggests that NeuralCut learns to exploit application-specific patterns to select good cuts.

\begin{table*}[t]
\caption{Training NeuralCut on a single domain (row) does not generalize well to other domains (columns). On all four benchmarks, the model that was trained on the same domain as it is tested on (i.e., main diagonal) performs best. Models that were trained on another domain when they are tested on perform significantly worse. The performance drop tends to be less severe when domains are more similar, e.g., packing and binary packing.}
\label{tab:transfer}
\begin{center}
\begin{small}
\begin{tabular}{lccccc}
\toprule
 && \multicolumn{4}{c}{\textbf{Reversed IGC integral ($\downarrow$) on test instances}, mean (ste)} \\
\noalign{\smallskip}\cline{3-6}\noalign{\smallskip}
 &&  \sc{Max. Cut} & \sc{Packing} & \sc{Bin. Packing} & \sc{Planning}  \\
\midrule
{\sc{Max. Cut}}        
&& \textbf{15.55 (0.09)} & 28.96 (0.03) & 26.33 (0.16) & 17.07 (0.07)\\
{\sc{Packing}}         
&& 25.03 (0.06) & \textbf{26.30 (0.08)} & 16.29 (0.27) & 29.67 (0.01)\\
{\sc{Bin. Packing}}      
&& 21.75 (0.08) & 27.66 (0.06) & \textbf{10.96 (0.33)} & 25.01 (0.12)\\
{\sc{Planning}}          
&& 19.13 (0.08) & 28.98 (0.03) & 22.83 (0.27) & \textbf{10.42 (0.04)}\\ 
\bottomrule	
\end{tabular}
\end{small}
\end{center}
\end{table*}
\subsection{Model Ablation}
\label{app:results:modelablation}
In Table \ref{tab:ablation-model-full}, we perform model ablations to better assess the effectiveness of our modelling choices. We begin with a simple GNN model that is based on \citet{GasseCFCL19}, but adapted to cut selection by using cuts in place of constraints in the original model formulation of \citet{GasseCFCL19} and switching the order in which half-convolutions are applied. Second, we use the features of NeuralCut ($+ \text{ our features}$) for both cuts and variables instead of the features of \citet{GasseCFCL19}. Third, we additionally make architectural choices, i.e. we replace layer normalization with batch normalization and add the attention module to the model. Finally, we leverage the tri-partite graph formulation to model constraints explicitly in addition to the cuts and variables, which gives the full NeuralCut model. 

\begin{table*}[t]
\caption{Our model choices tend to improve both bound fulfillment (left) and test reversed IGC integral (right) on all four benchmarks. The improvements tend to be clearer for bound fulfillment and binary packing. \cite{GasseCFCL19} is the model and features described in \citep{GasseCFCL19} and adapted to cut selection by replacing constraints with cuts in the bi-partite graph. For the purpose of this ablation, this base model is first augmented with the features of NeuralCut. Second, structural changes are made to the model by adding the attention module and using batch normalization. Finally, constraints are explicitly modelled in the tripartite graph giving the full NeuralCut model.}
\label{tab:ablation-model-full}
\begin{center}
\begin{small}
\adjustbox{max width=\textwidth}{%
\begin{tabular}{lcccccccccc}
\toprule
 && \multicolumn{4}{c}{\textbf{Bound fulfillment ($\uparrow$) on test samples}, mean} & & \multicolumn{4}{c}{\textbf{Reversed IGC integral ($\downarrow$) on test instances}, mean (ste)} \\
\noalign{\smallskip}\cline{3-6}\cline{8-11}\noalign{\smallskip}
 && \sc{\scriptsize Max. Cut} & \sc{\scriptsize Packing} & \sc{\scriptsize Bin. Packing} & \sc{\scriptsize Planning} & &  \sc{\scriptsize Max. Cut} & \sc{\scriptsize Packing} & \sc{\scriptsize Bin. Packing} & \sc{\scriptsize Planning}  \\
 \midrule
{\citet{GasseCFCL19} }
&& 0.90 & 0.42 & 0.54 & 0.95     
&&  16.68 (0.10) & 27.30 (0.07) & 16.09 (0.31) & 11.26 (0.06) \\
{ $+ \text{ our features}$}
&& 0.93 & 0.26 & 0.66 & \textbf{0.99}
&& \textbf{15.54 (0.09)} & 28.54 (0.04) & 12.77 (0.32) & 10.66 (0.04)\\
{ $+ \text{ our architecture}$}
&& 0.93 & \textbf{0.62} & 0.76 & \textbf{1.00}
&& \textbf{15.72 (0.09)} & \textbf{26.44 (0.07)} & \textbf{10.73 (0.33)} & 10.55 (0.04) \\
\midrule[0.2pt]
{NeuralCut ($+ \text{ our graph}$)}        
&& \textbf{0.96} & \textbf{0.61} & \textbf{0.78} & \textbf{1.00}           
&& \textbf{15.55 (0.09)} & \textbf{26.30 (0.08)} & \textbf{10.96 (0.33)} & \textbf{10.42 (0.04)}\\
\bottomrule	
\end{tabular}
} 
\end{small}
\end{center}
\vskip -0.1in
\end{table*}
\subsection{Loss Ablation}
\label{app:results:lossablation}
In Table \ref{tab:ablation-full-loss}, we experimented with different loss functions on the bound fulfillment surrogate. In addition to the binary entropy loss (which was used to train all NeuralCut models), we explored the effectiveness of mean squared error (MSE), both with linear and sigmoid activations as well as a cross entropy loss function. 

\begin{table*}[t]
\caption{Different choices for the training objective (where bound fulfillment is used as surrogate) tend to produce models with comparable test performance. Notable exception are binary packing and the choice of cross entropy for packing. Binary entropy (as was used to train NeuralCut) tends to be a good choice across all four benchmarks. Best are bold-faced.}
\label{tab:ablation-full-loss}
\begin{center}
\begin{small}
\adjustbox{max width=\textwidth}{%
\begin{tabular}{lcccccccccc}
\toprule
 && \multicolumn{4}{c}{\textbf{Bound fulfillment ($\uparrow$) on test samples}, mean} & & \multicolumn{4}{c}{\textbf{Reversed IGC integral ($\downarrow$) on test instances}, mean (ste)} \\
\noalign{\smallskip}\cline{3-6}\cline{8-11}\noalign{\smallskip}
 && \sc{\scriptsize Max. Cut} & \sc{\scriptsize Packing} & \sc{\scriptsize Bin. Packing} & \sc{\scriptsize Planning} & &  \sc{\scriptsize Max. Cut} & \sc{\scriptsize Packing} & \sc{\scriptsize Bin. Packing} & \sc{\scriptsize Planning}  \\
 \midrule
 {Binary Ent. (NeuralCut) }        
&& \textbf{0.96} & \textbf{0.61} & \textbf{0.78} & \textbf{1.00}           
&& \textbf{15.55 (0.09)} & \textbf{26.30 (0.08)} & \textbf{10.96 (0.33)} & \textbf{10.42 (0.04)}\\
 \midrule[0.2pt]
{MSE (linear act.)}
&& \textbf{0.96} & \textbf{0.60} & 0.70 & \textbf{0.99}    
&&  \textbf{15.59 (0.09)} & \textbf{26.37 (0.08)} & 11.66 (0.33) & 10.54 (0.05) \\
{MSE (sigmoid act.)}
&& \textbf{0.97} & \textbf{0.61} & 0.69 & \textbf{1.00}
&& \textbf{15.63 (0.09)} & \textbf{26.30 (0.08)} & 12.65 (0.32) & \textbf{10.47 (0.04)}\\
{Cross Entropy}
&& \textbf{0.96} & {0.30} & 0.73 & \textbf{1.00}
&& \textbf{15.71 (0.09)} & 28.73 (0.04) & \textbf{11.36 (0.34)} & 10.59 (0.04) \\
\bottomrule	
\end{tabular}
} 
\end{small}
\end{center}
\vskip -0.1in
\end{table*}

\section{Q\&A Discussion}\label{app:qa}

\paragraph{Would NeuralCut generalize to different test problems without the need of re-training? What about heterogeneous instances such as those in MIPLIB?}
{
The MIPLIB dataset has been developed to benchmark general-purpose solvers, and thus contains diverse instances that vary in size, complexity and structure. 
ML for MILP may be most promising when used to learn application-specific models that can exploit structural patterns between problem instances arising from a common distribution \cite{BengioLP21}. Indeed, this is the setting on which many recent works in ML for MILP operate\footnote{See also the recent NeurIPS competition \url{https://www.ecole.ai/2021/ml4co-competition/\#datasets}.
} and highly relevant in practice. Therefore, even though we demonstrate the effectiveness of \emph{Lookahead} across heterogenous problems (Figure~\ref{fig:justification}), we did not focus on learning models for MIPLIB but trained application-specific ones instead. 
Regarding generalization, we observed that training on a single domain is unlikely to generalize to other domains. In additional evaluations, we assess how the models trained on the datasets from \citet{TangAF20} perform when used on families different from the one on which they were trained on, and found that performance dropped significantly. This suggests that NeuralCut learns to exploit application-specific patterns to select good cuts.
}

\paragraph{What are the costs of deploying NeuralCut in a realistic MILP solver?}
{
While previous work evaluated their models within a naive stripped-down B\&C procedure on synthetic instances only, we directly embed our models into SCIP and subject them to the complexity of operating with all the other components of the solver, demonstrating improvements on real-world instances from NN verification. 
We believe our work is a push forward and that our results are encouraging, as they highlight the potential of NeuralCut to improve MILP solver performance in specific applications.
At the same time, it is true that any ML method may suffer from ``burn-in cost'' for training and potential overhead in deployment. In our setting, the cost of training mostly comes down to the computation of the LA expert decisions. Once trained, though, a single learned policy can be applied to several MILP instances of the same family, effectively amortizing this labeling cost. At deployment, the computational overhead comes from extracting features and performing a forward-pass through the model on CPU. 
Finally, note that for some difficult problems spending extra time and computational budget at early stages of B\&C might payoff in later optimization phases. 
}

\paragraph{Why selecting only one cut at a time? Could you look ahead further then one step?}
{
Directly inspired by cutting plane selection in the solver, our model selects a single cut at a time, but it is used to select multiple cuts from the same cutpool by iterative application. 
Indeed, one could define a lookahead score for $k$-subsets of cuts (to select multiple cuts at once) or based on a $k$-step lookahead window (i.e., computing explicit bound improvements after $k$ cut additions). We choose not to, because the cost of computing such a score grows exponentially in $k$; for this reason, it is also intractable to compute guarantees. 
While the idea of simultaneously selecting multiple cuts relates to techniques like ranking and multiple instance learning \citep[see, e.g.,][]{HuangEtAl21}, the theme of multi-step lookahead has been examined for the task of variable selection in B\&B, e.g., by \citet{GlankwamdeeL11,Schubert17}.
}

\end{document}